\documentclass[letterpaper]{article} 
\usepackage{aaai25}  
\usepackage{times}  
\usepackage{helvet}  
\usepackage{courier}  
\usepackage[hyphens]{url}  
\usepackage{graphicx} 
\usepackage{natbib}  
\usepackage{caption} 
\usepackage{xcolor}
\usepackage{amsmath}
\DeclareMathOperator*{\argmax}{arg\,max}

\DeclareMathOperator*{\argtopk}{\mathop{\mathrm{arg\,top}\,k}}

\usepackage{algorithm}
\usepackage{algorithmic}
\usepackage{tabularx} 
\usepackage{booktabs} 
\usepackage{multirow} 
\usepackage{amsmath}
\usepackage{graphicx} 
\usepackage{subcaption} 
\usepackage{newfloat}
\usepackage{listings}
\DeclareCaptionStyle{ruled}{labelfont=normalfont,labelsep=colon,strut=off} 
\floatstyle{ruled}
\newfloat{listing}{tb}{lst}{}
\floatname{listing}{Listing}
\title{Towards Understanding Human Emotional Fluctuations \\ with Sparse Check-In Data}
\author{
    Sagar Paresh Shah\textsuperscript{\rm 1}\textsuperscript{\rm 2},
    Ga Wu\textsuperscript{\rm 1},
    Sean W. Kortschot\textsuperscript{\rm 2},
    Samuel Daviau\textsuperscript{\rm 2}
}
\affiliations{
    \textsuperscript{\rm 1}Faculty of Computer Science, Dalhousie University, Halifax, Canada\\
    \textsuperscript{\rm 2}UpBeing Inc., Halifax, Canada\\
    sg355741@dal.ca,
    ga.wu@dal.ca,
    sean@upbeing.ai,
    sam@upbeing.ai
}

\begin{document}

\maketitle

\begin{abstract}
Data sparsity is a key challenge limiting the power of AI tools across various domains. The problem is especially pronounced in domains that require active user input rather than measurements derived from automated sensors. It is a critical barrier to harnessing the full potential of AI in domains requiring active user engagement, such as self-reported mood check-ins, where capturing a continuous picture of emotional states is essential. In this context, sparse data can hinder efforts to capture the nuances of individual emotional experiences such as causes, triggers, and contributing factors. 
Existing methods for addressing data scarcity often rely on heuristics or large established datasets, favoring deep learning models that lack adaptability to new domains. This paper proposes a novel probabilistic framework that integrates user-centric feedback-based learning, allowing for personalized predictions despite limited data. Achieving 60\% accuracy in predicting user states among 64 options (chance of 1/64), this framework effectively mitigates data sparsity. It is versatile across various applications, bridging the gap between theoretical AI research and practical deployment.
\end{abstract}

\section{Introduction}

Human emotions are complex and multifaceted. Early research by \citeauthor{picard2000affective} and \citeauthor{emotion_recognition_hci} demonstrated the potential of technology to understand and influence emotions. In particular, emotion prediction can enable timely interventions for negative emotions like stress or anxiety \cite{de2013predicting}. It aids in aligning tasks with periods of peak focus \cite{sun2023challenges} and offering personalized recommendations \cite{emotion_recognition_music_listening_article}. 

\begin{figure}
    \centering
    \includegraphics[width=\linewidth]{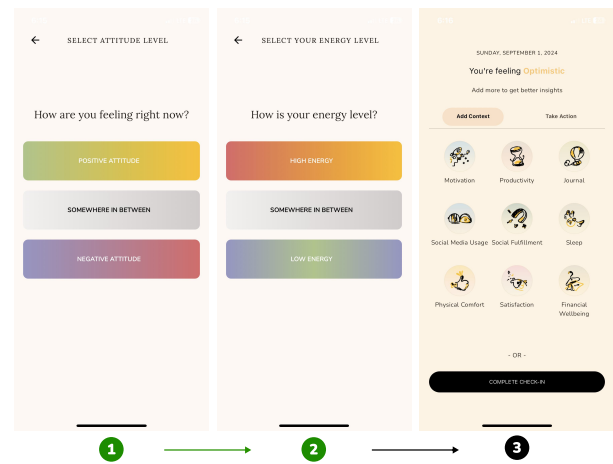}
    \caption{User interface of UpBeing App. 1)-2), the user emotion check-in interface. 3) the homepage of UpBeing where user selected emotion is presented to the user.}
    \label{fig:upbeing_app}
\end{figure}

Accurately predicted human emotions at points in time where their emotional state is unknown holds significant potential for building empathetic A.I. (EAI) systems. EAI is a set of AI systems that are capable of adaptively understanding and responding to an end user's emotional state \cite{srinivasan2022roleofempathy}.
However, achieving this goal presents a significant challenge: data sparsity. The challenge of data sparsity, especially in domains reliant on user-generated input, poses a significant obstacle to the effective deployment of AI technologies. Traditional AI approaches often require large datasets or rely on heuristic-driven models that lack adaptability to domain-specific, data-scarce environments. To create truly empathetic and responsive systems, understanding a user's emotional nuances (with well-explored psychology knowledge) is essential for designing AI tools that offer tailored experiences and support \cite{de2013predicting} \cite{srinivasan2022roleofempathy}.

This paper proposes Mood Shift Predictor with Sparse Check-ins (MSPSC), a novel emotion recognition system that integrates the psychological understanding of human emotion shift and advanced machine learning techniques. Compared with handy solutions that naively adopt existing general-purpose machine learning models, MSPSC can adapt to an individual's emotional patterns over time without requiring a large amount of initial data. In particular, the proposed method dynamically compares relevant input parameters, effectively alleviating the data sparsity problem and avoiding the limitations of traditional models that rely on a fixed set of features. Our empirical studies show that MSPSC significantly improves the emotion prediction performance over existing models and demonstrates the potential to support other human-centric machine learning applications where data scarcity poses significant challenges.

\section{Preliminary and Background}
The landing application of this research is an adaptive emotion prediction system, namely UpBeing, driven by sparse human check-in data. By understanding the dynamics of \textit{core emotion} \cite{core_affect_doi:10.1080/02699930902809375}, the AI system can respond adaptively to the user's emotional state. For example, if a fitness tracker shows low activity levels and the weather is gloomy, the system can suggest uplifting activities or content to improve the user's mood, creating a more personalized experience \cite{12_point_core_affect_article}. The concept of core emotion is also fundamental to predicting emotional responses. By assessing an individual's core effect on a specific stimulus, it is possible to predict their emotional response in similar situations in the future \cite{core_affect_doi:10.1080/02699930902809375}.

\subsection{Existing Emotion Shift Prediction Approaches}

Accurately predicting mood shifts is challenging due to the complexity of human emotions, personal experiences, and the dynamic nature of emotional states \cite{barrett2017theory}. In their study, \citeauthor{van2016exploring} assessed various machine learning models, including ARIMA, DTW, SVM, and Random Forests, for predicting short-term mood changes in individuals with depression using phone sensor data. While these models showed some improvements, they were limited by data sparsity and the sensor data's relevance to short-term mood changes, highlighting the need for more sophisticated models and relevant data in e-health applications \cite{van2016exploring}. Similarly, \citeauthor{shi2018machine} explored mood prediction among medical interns using mobile health data and found that despite the complexity of neural networks, their accuracy was constrained by high rates of missing data \cite{shi2018machine}, aligning with our observations on the impact of sparse data on mood prediction.

\subsubsection{Machine Learning Methodologies:}
Machine learning is crucial for analyzing user data and predicting emotional states, initially focusing on physiological signals like heart rate variability \cite{emotion_recognition_music_listening_article}. More recent methods use user-generated content, such as text and interaction data, to predict emotions \cite{tato2018predicting} \cite{abburi2017multimodal} \cite{guo2019personalized}. \citeauthor{bai2022training} examined reinforcement learning with human feedback (RLHF) for aligning language models with helpful and unbiased behavior. Reinforcement learning differs from supervised learning by actively exploring its environment \cite{kaelbling1996reinforcementlearningsurvey}, which helps avoid biases from frequently reported emotions, thus improving prediction accuracy. Human-labeled data is vital for training emotion models, as shown by \citeauthor{article_body_expressions} and \citeauthor{Moerland2017EmotionIR}, who highlighted the role of human expertise and emotional feedback in improving model performance \cite{article_ml_emotion_comprehensive_review}. \citeauthor{article_online_learning} also proposed reinforcement learning for emotion prediction using physiological signals.

Recent observations highlight the potential of machine learning for mood shift prediction, utilizing diverse data sources like user-generated content, physiological signals, and app usage patterns for a comprehensive view of emotional states \cite{tato2018predicting}. Deep learning advancements address sparse data challenges by detecting intricate patterns and continuously improving accuracy through real-time data and user feedback \cite{deep_learning_article} \cite{settles2009active}.

\subsubsection{Psychological Understanding of Human Moods:}
{\textit{Core affect}} is a well-established concept that maps human emotions into a Euclidean space \cite{core_affect_doi:10.1080/02699930902809375}, using valence charts to break down emotions into two dimensions: attitude (displeasure-pleasure) and activation (low energy-high energy) \cite{12_point_core_affect_article}. For example, \textit{serene} (positive attitude, low activation) and \textit{furious} (negative attitude, high activation) are plotted based on these dimensions. Despite limitations like cultural variations and the imprecision of mapping emotions at regular intervals \cite{12_point_core_affect_article}, this framework formed the foundation of the emotion grid in the UpBeing App.

\begin{figure}
    \centering
    \includegraphics[width=\linewidth]{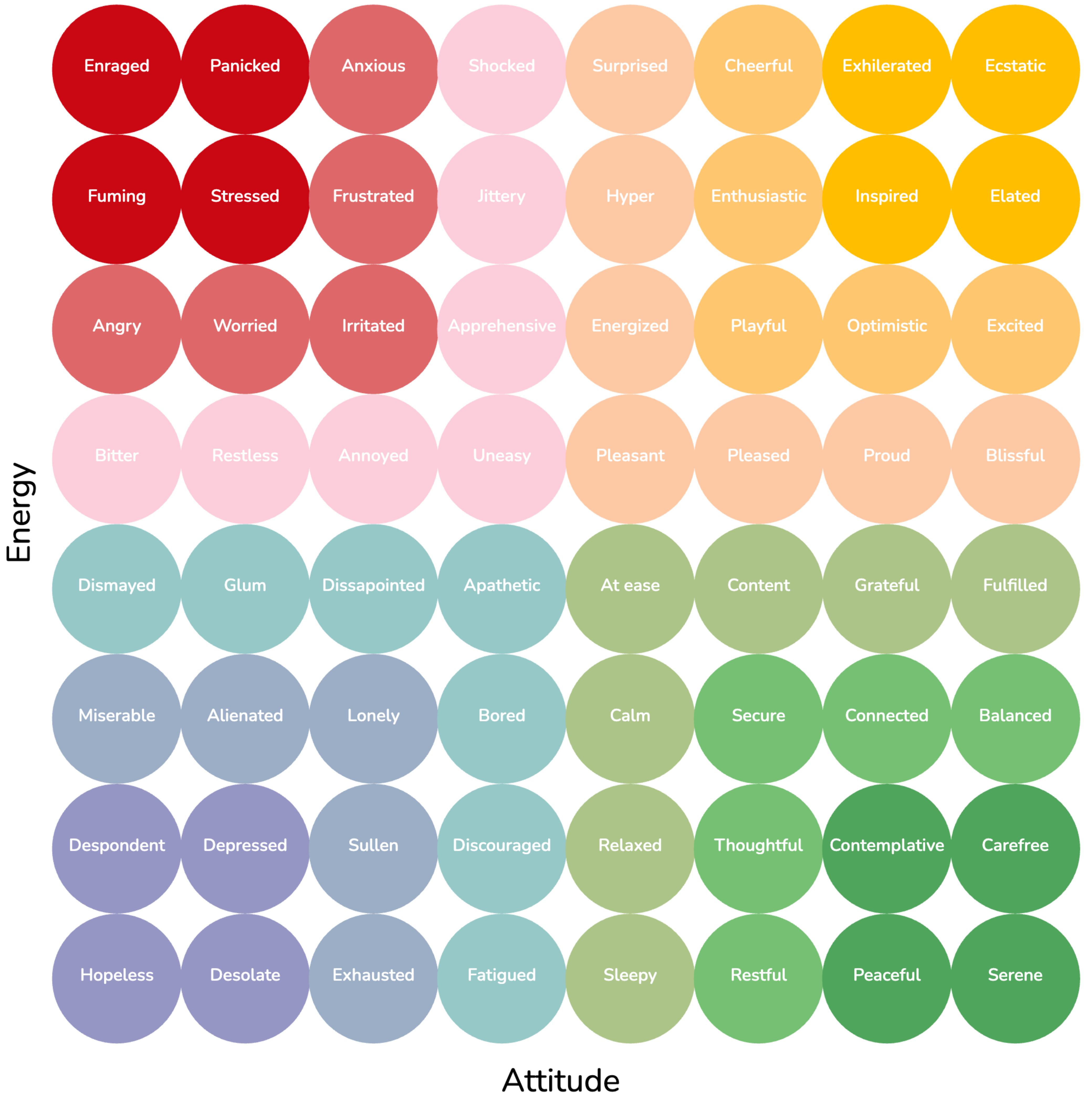}
    \caption{Emotion grid that reflects core affect. Attitude axis goes from left to right, indicating negative to positive. Energy axis goes from bottom to top, indicating high to low.}
    \label{fig:emotion_grid}
\end{figure}

\begin{figure*}
    \centering
    \includegraphics[width=\linewidth]{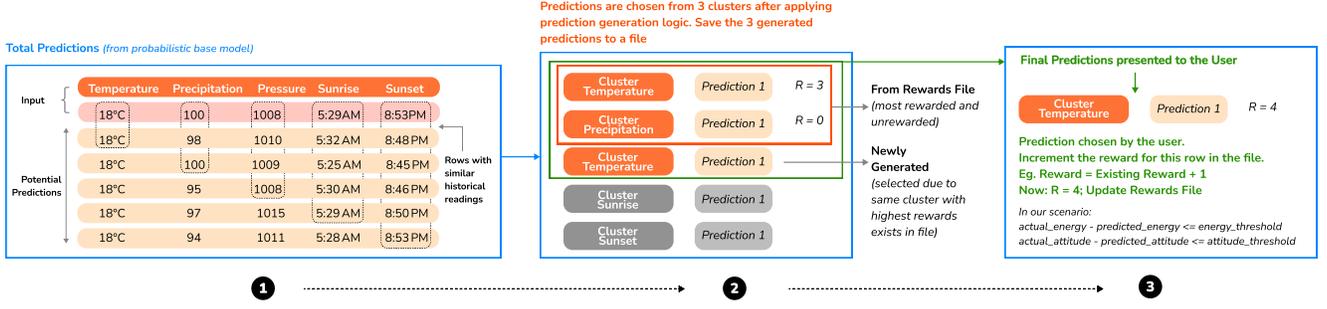}
    \caption{Overview of the proposed emotion prediction system. 1) Relevant Historical Check-ins Retrieval, 2) Personalized Environmental Influence with Utilities, and 3) Personalized Emotion Prediction.}
    \label{fig:overview-proposed-solution}
\end{figure*}

\section{Mood Shift Predictor with \\ Sparse Check-ins}

We aim to develop a personalized emotion prediction model that models individual emotional status, $e^u \in \mathcal{E}$ by taking sparse user check-in history $H^u = \{\cdots \mathbf{h}_t^u \cdots\}$ along with limited user-granted side-information accesses, $\mathbf{s}^u$. 

In particular, the user check-in history $H^u$ is sparse and the check-in is a combination of a single mood selection and corresponding user-granted side information. The single mood choice is over a pre-defined emotion grid among 64 options, denoted as $\mathcal{E}$, that aligns with the core effect of psychological understanding as shown in Figure~\ref{fig:emotion_grid}. We call the user check-in history as {\it active data sources}. Figure \ref{fig:upbeing_app} demonstrates the user process of completing a check-in with simple steps within the mobile app.

For user-granted side information (a.k.a environmental factors), there are three major groups of data sources based on the level of users' consent on their privacy as follows.

\begin{enumerate}
    \item{\textbf{Weather Information}}: Access to weather data through linked apps can reveal correlations between conditions (e.g., sunny mornings) and mood, with individuals reacting differently to weather, as \citeauthor{klimstra2011come} noted.
    \item{\textbf{Calendar Events}}: Linked calendar data helps identify mood triggers related to schedules, such as a busy day correlating with feelings of Tiredness or Overwhelm, consistent with \citeauthor{larsen1990individual} findings on weekly mood fluctuations.
    \item{\textbf{Fitness Tracker Data}}: Provide detailed data on physical activity and well-being, offering insights into mood prediction, though adoption is limited by privacy concerns, as supported by the research of \citeauthor{8925520_garmin_fitness}. 
\end{enumerate}

The distribution of data sources reveals significant sparsity, with weather data comprising 35.4\% of the dataset due to its public availability. In contrast, data that includes both check-ins and calendar events account for only 10.5\%, and data from wearables (Terra) is as low as 1.3\%. This highlights the challenges posed by limited access to comprehensive data sources, contributing to the overall issue of data sparsity in mood prediction research and leading existing predictive models inapplicable.

\subsection{Probabilistic Assumption of Emotion Prediction}
We assume that human emotion can be estimated through a simplified probabilistic function in the form of 
\begin{equation}
    P(\mathcal{E}^u| H^u, \mathbf{s}^u, \mathbf{w}^u).
\label{eq:init}
\end{equation}
Intuitively, the probabilistic function attributes an individual's emotional status to three factors: 1) his/her historical emotional status $H^{u}$, 2) current environmental state of the individual $\mathbf{s}^u$, and 3) the individual's inherent mood tendency $\mathbf{w}^u$.

As the joint effect of the three factors is unknown, we further introduce a hidden variable vector $\mathbf{z} \in Z$ to model the joint effect in a factorized way. Formally, we derive Equation~\ref{eq:init} with the introduced hidden variable as follows:
\begin{equation}
\begin{aligned}
    P(\mathcal{E}^u| H^u, \mathbf{s}^u, \mathbf{w}^u) & = \sum_k P(\mathcal{E}^u, \mathbf{z}^k| H^u, \mathbf{s}^u, \mathbf{w}^u)\\
    & = \sum_k \prod_{t=1}^T P(\mathcal{E}^u, \mathbf{z}^k| \mathbf{h}_t^u, \mathbf{s}^u, \mathbf{w}^u)\\
    & = \sum_k \prod_{t=1}^T\prod_{i=1}^M P(\mathcal{E}^u, z_i^k| \mathbf{h}_t^u, s_i^u, \mathbf{w}^u),\\
\end{aligned}
\end{equation}
where we decomposed the user's historical check-ins $H^u$ into independent contributors to the mood estimation task. Similarly, we also decomposed the environmental information $\mathbf{s}^u$ into independent environmental factors.

Now, considering each environmental factor may affect an individual's emotion constantly and independently (e.g. raining days may evoke sadness for certain individuals always), we treat hidden variable vector $\mathbf{z}$ as the outcome of deterministic similarity measurement between current environmental facts and historical moments the user experienced. Specifically,
\begin{equation}
\begin{aligned}
&\sum_k \prod_{t=1}^T\prod_{i=1}^M P(\mathcal{E}^u, z_i^k| \mathbf{h}_t^u, s_i^u, \mathbf{w}^u) \\ 
& = \sum_k \prod_{t=1}^T\prod_{i=1}^M P(\mathcal{E}^u| z_i^k, \mathbf{w}^u) P (z_i^k|\mathbf{h}_t^u, s_i^u)\\
& = \prod_{t=1}^T \prod_{i=1}^M \underbrace{P(\mathcal{E}^u| z_i, \mathbf{h}_t^u, \mathbf{w}^u)}_{\substack{\text{Personalized} \\ \text{Environmental Influence}}} \underbrace{\delta [ z_i - \phi(\mathbf{h}_t^u, s_i^u)]}_{\substack{\text{Per-environmental Factor}\\\text{Similarity Measure}}}, \\
\end{aligned}
\end{equation}
where summation over hidden variable $\mathbf{z}$ is removed due to the deterministic nature of $P (z_i^k|\mathbf{h}_t^u, s_i^u)$ (as denoted as a delta function). $\phi$ denotes a similarity function.

To prevent numerical issues, we use the logarithm form of the above equation in practice, such that 
\begin{equation}
    \sum_{t=1}^T \sum_{i=1}^M 
    \log P(\mathcal{E}^u| z_i, \mathbf{h}_t^u, \mathbf{w}^u) \delta [ z_i - \phi(\mathbf{h}_t^u, s_i^u)]
\label{eq:log_obj}
\end{equation}

The above derivation revealed a logical path of probabilistically modeling human emotion: Searching for historically similar circumstances of the individual (under each passive data factor) and weighted historical user mood check-ins as predictions in a personalized manner. The personalization step reflects the reality that some people are easily influenced by weather, whereas others are not.

\subsection{Relevant Historical Check-ins Retrieval}

Now, we describe how to measure the similarity between an environmental factor $s_i^u$ and a historical check-in $\mathbf{h}_t^u$. 

Since historical check-in $\mathbf{h}_t^u$ contains both user self-reported mood choice $e_t^u$ and corresponding environmental information $\mathbf{s}_t^u$ at time $t$ as a tuple $\mathbf{h}_t^u = (e_t^u, \mathbf{s}_t^u)$, the similarity measurement of $ \phi(\mathbf{h}_t^u, s_i^u)$ can be reduced to measuring the difference of the environmental factor only as follows:

\begin{equation}
    z_i = \phi(\mathbf{h}_t^u, s_i^u) = \frac{1}{|s_{t,i}^u - s_i^u|}/Z,
\label{eq:similarity}
\end{equation}
where $s_{t,i}^u$ is one element of history $\mathbf{s}_{t}^u$ that reflects the environmental factor $i$. And, $Z$ is the partition function that keeps the similarity measurement in a valid range ($[0,1]$) by enforcing $\sum_t \phi(\mathbf{h}_t^u, s_i^u) = 1$.

It is worth noting that historical check-ins may not contain specific environmental factors $s_{t,i}^u$ due to various reasons, such as temporarily disabled user consents, and unexpected service outages. To handle those cases, we set similarity 
\begin{equation}
    \phi(\mathbf{h}_t^u, s_i^u) = 0 \quad \text{if } s_{t,i}^u = \text{ N/A}
\end{equation}

The similarity score is used to determine how well a current environmental factor aligns with historical data, aiding in historical emotional states search, thereby identifying the closest match given the similar proxy conditions. Intuitively, people who have experienced similar external environments in the past are likely to evoke similar emotional states in the future when the environmental factors match, so searching/ranking similar external environments can support emotion prediction.

\subsection{Personalized Environmental Influence}

Given relevant historical check-ins retrieved from the previous step, one can simply define an environmental influence function
$P(\mathcal{E}^u| z_i, \mathbf{h}_t^u, \mathbf{w}^u)$ to complete the emotion prediction by conducting influence aggregation that
\begin{equation}
    P(\mathcal{E}^u = e \mid z_i, \mathbf{h}_t^u, \mathbf{w}^u) = z_i \cdot \mathbf{1}[e_t^u = e] \quad \forall e \in \mathcal{E}
    \label{eq:simple_environmental},
\end{equation}
where personalization factors $
\mathbf{w}^u$ is completely ignored.

By combining Equation~\ref{eq:log_obj},~\ref{eq:similarity}, and \ref{eq:simple_environmental}, we are able to predict users' mood status distribution (spanning over 64 categorical choices) with sparse check-in records.
 
{\bf Limitation of the simple solution:} The simple solution is helpful, but it does not fully incorporate the psychological expectation of emotion prediction in practice. Psychologist Barry Schwartz emphasizes that while some choices are beneficial, an abundance of options can lead to decision paralysis, known as the paradox of choice \cite{schwartz2015paradox}. This phenomenon, supported by Hick's Law, suggests that decision-making time increases with the number of options, potentially causing confusion \cite{proctor2018hick}. In the context of emotional predictions, this paradox becomes evident when the simple model generates numerous potential predictions based on similarity matching between current inputs and historical data. While this approach may increase the number of options, it also risks overwhelming the user, as predicted by Hick's Law.

To address the over-smoothed prediction problem above, instead of working with Equation~\ref{eq:simple_environmental}, we consider including personalization in the mood prediction task. Specifically, we define the environmental influence function by leveraging variable vector $\mathbf{w}^u$, such that
\begin{equation}
    P(\mathcal{E}^u = e \mid z_i, \mathbf{h}_t^u, \mathbf{w}^u) = z_i \cdot w_i^u \cdot \mathbf{1}[e_t^u = e] \quad \forall e \in \mathcal{E}
    \label{eq:environmental},
\end{equation}
where 
$\mathbf{w}^u \in \mathbf{R}^M$
aligns the dimension of environmental factors. Intuitively, the personalization coefficient reflects an individual's tendency to be affected by certain external environmental factors. Indeed, some might be influenced by weather, whereas others are not. Note, the personalization coefficient $\mathbf{w}^u$ is not observable in historical data, which needs further estimation to obtain.

\subsubsection{Cluster of Environmental Influence:} 
To reveal a user's tendency to be influenced by certain external factors, we introduce a cluster of external factors for each user. Specifically, for each environmental factor \( s_i^u \) (e.g., temperature, precipitation), we gather a small group of historical check-ins \( \mathbf{h}_{t,j}^u \) using the similarity score \( \phi(\mathbf{h}_t^u, s_i^u) \) with top-k selection. Formally, clusters \( \mathcal{C}_i \) are defined as follows:

\begin{equation}
\mathcal{C}_i = \left\{ \mathbf{h}_{t}^u \mid  \argtopk_{\mathbf{h}_t^u} \phi(\mathbf{h}_t^u, s_i^u) ~~ \forall \mathbf{h}_{t}^u \in H^u \right\}.
\end{equation}
With the cluster definition, the remaining task is simply to look for which clusters the user might be affected most. To this end, we introduce a utility function from interactive recommendation literature~\cite{huang2011designing}, where we model users' personalized environmental influences by updating utilities.

\begin{table*}[t]
    \centering
    \setlength\extrarowheight{-3pt}
    \resizebox{\linewidth}{!}{
    \begin{tabularx}{\textwidth}
        {p{2.7cm}|>{\raggedleft\arraybackslash}p{2cm}|>{\raggedleft\arraybackslash}X|>{\raggedleft\arraybackslash}p{2cm}|>{\raggedleft\arraybackslash}X}
        \toprule
        
        \multirow{2}{*}{\textbf{Model}}&\multicolumn{2}{c|}{\textbf{Consistent Users}}&\multicolumn{2}{c}{\textbf{Inconsistent Users}}\\
        \cmidrule(lr){2-5}
        & Total Rows & \% Avg. Correct Predictions $\uparrow$ & Total Rows & \% Avg. Correct Predictions $\uparrow$\\
        \midrule

        Linear Regression & 4111 & 7.39 & 2366 & 5.23 \\
        KNN & 4111 & 21.59 & 2366 & 24.08 \\
        XGBoost & 4111 & 27.66 & 2366 & 30.78 \\
        Transformer & 4111 & 7.07 & 2366 & 2.78 \\
        Autoencoder & 4111 & 14.47 & 2366 & 16.19 \\
        
        \textbf{MSPSC} & \textbf{4110} & \textbf{46.47} & \textbf{2366} & \textbf{48.35} \\

        \bottomrule
    \end{tabularx}}
    \caption{Quantitative evaluation on 5 test sets. Models trained on a holdout training dataset. Performance assessed for consistent and inconsistent users. The up arrow indicates higher is preferable and the down arrow indicates lower is preferable.}
    \label{tab:comparative_analysis}
\end{table*}

\begin{table*}[t]
    \centering
    \setlength\extrarowheight{-3pt}
    \resizebox{\linewidth}{!}{
    \begin{tabularx}{\textwidth}
        {p{2.7cm}|>{\raggedleft\arraybackslash}p{1.4cm}|>{\raggedleft\arraybackslash}p{1.3cm}|>{\raggedleft\arraybackslash}p{1.7cm}|>{\raggedleft\arraybackslash}X|>{\raggedleft\arraybackslash}X|>{\raggedleft\arraybackslash}p{1.4cm}|>{\raggedleft\arraybackslash}p{1.5cm}}
        \toprule

        \textbf{Model} & \textbf{Total Training Rows} & \textbf{Total Test Rows} & \textbf{\% Correct Predictions $\uparrow$} & \textbf{No. of Correct Predictions $\uparrow$} & \textbf{Avg. Predictions Per Test Row} & \textbf{Avg. Delta (Energy)$\downarrow$} & \textbf{Avg. Delta (Attitude)$\downarrow$} \\
        \midrule
        
        Linear Regression & 757 & 318 & 3.14 & 10 & 1 & 32.77 & 51.5\\ 
        KNN & 757 & 318 & 45.59 & 145 & 1 & 19.71 & 13.44 \\ 
        XGBoost & 757 & 318 & 21.38 & 68 & 1 & 20.26 & 17.46 \\ 
        Autoencoder & 757 & 318 & 6.6 & 21 & 1 & 22.21 & 22.16 \\ 
        Transformer & 757 & 318 & 9.12 & 29 & 1 & 21.28 & 17.07 \\ 

        \textbf{MSPSC} & \textbf{757} & \textbf{318} & \textbf{64.15} & \textbf{204} & \textbf{2.55} & \textbf{18.83} & \textbf{23.77} \\
            
        \bottomrule
    \end{tabularx}}
    \caption{Case study on a typical heavy user. Models are trained on a unified dataset and tested on an individual's test set. Delta indicates the prediction error. The up arrow indicates higher is preferable and the down arrow indicates lower is preferable}
    \label{tab:comparative_analysis_single_user}
\end{table*}

\subsubsection{Utility Function:} 
We define the utility function on top of the clusters that were created in the previous step, where we reward clusters based on how well predictions align with user-actively reported emotions over time. Specifically, let \( \hat{e}_i^u \) represent a major vote prediction over historical check-ins in cluster $\mathcal{C}_i$ 
and \( {e}^u \) be the user's self-reported emotion (ground-truth check-in). The utility function \( R_u(C_i) \) updates user's personalized environmental influence $\mathbf{w}^u$ as:
\begin{equation}
w_i^u \leftarrow R(C_i, w_i^u) = \begin{cases} 
 w_i^u +1 & \text{if } [| \hat{e}_i^u - {e}^u |] \leq \epsilon \\
0 & \text{otherwise},
\end{cases}
\label{eq:reward_function}
\end{equation}
where \( [| \hat{e}_i^u - {e}^u |] \) measures the deviation of the predicted emotion \(\hat{e}_i^u \) of cluster $\mathcal{C}_i$ with the user self-reported emotion \( e^u \) on the X and Y axes of emotion grid (Figure~\ref{fig:emotion_grid}), with tolerance $\epsilon$. The tolerance threshold is a hyper-parameter that helps battle the data sparsity issue. In this particular application, we set it to $\epsilon = 13$. User feedback refines the reward system, improving the focus on impactful environmental factors.

Now, by combining Equation~\ref{eq:environmental} and the previously described similarity function (Equation ~\ref{eq:similarity}), we now are able to produce personalized emotion prediction with
\begin{equation}
    \argmax_{e\in\mathcal{E}}\sum_{t=1}^T \sum_{i=1}^M 
    \log P(e| z_i, \mathbf{h}_t^u, \mathbf{w}^u) \delta [ z_i - \phi(\mathbf{h}_t^u, s_i^u)]
\label{eq:final}
\end{equation}

Since the entire training and inference processes do not involve batch training on historical data, the proposed method, MSPSC, shows a significant advantage in supporting prediction with sparse check-in data. 

\section{Empirical Evaluation and Comparison}

In this section, we assess the performance of commonly used prediction models alongside our proposed MSPSC model. Users in our empirical study are categorized into "Most Consistent" (at least 2 daily check-ins for 7 days) and "Least Consistent" (at most 1 daily check-in, non-consecutive). Consistent users provide valuable data for identifying emotion patterns while in-consistent users make it challenging to establish patterns.

\subsubsection{Evaluation Metric:} We evaluate models' performance with accuracy that compares the predicted responses against actual user responses within specified tolerances for energy and attitude. The accuracy function identifies matching predictions (options presented to the user) that fall within these tolerances, marking them as accurate and adjusting the reward system accordingly. The \textit{Delta} marks average absolute difference between predicted and actual values.

\subsubsection{Baselines:}
We compare the proposed approach with the following three common supervised learning models for emotion prediction, including Linear Regression, K-Neighest Neighbors (KNN), and XGBoost
\cite{field2012discovering}  \cite{chen2016xgboost}. Additionally, we explored the unsupervised feature learning model, Auto-encoder \cite{rumelhart1986learning}, and sequential data model, Transformer \cite{vaswani2017attention}. The selection of baselines were conditioned on the algorithms' generalization ability while combating data sparsity. We do not include complex models (e.g. complex neural networks) in this work due to the data sparsity.

For neural network models (Auto-encoder and Transformer), the best hyper-parameter configurations are concluded as a batch size of 64, a lower learning rate of 0.0001, and 25 epochs.

\subsubsection{Results:}

\begin{figure*}
    \centering
    \begin{subfigure}{0.325\linewidth}
        \centering
        \includegraphics[width=\linewidth]{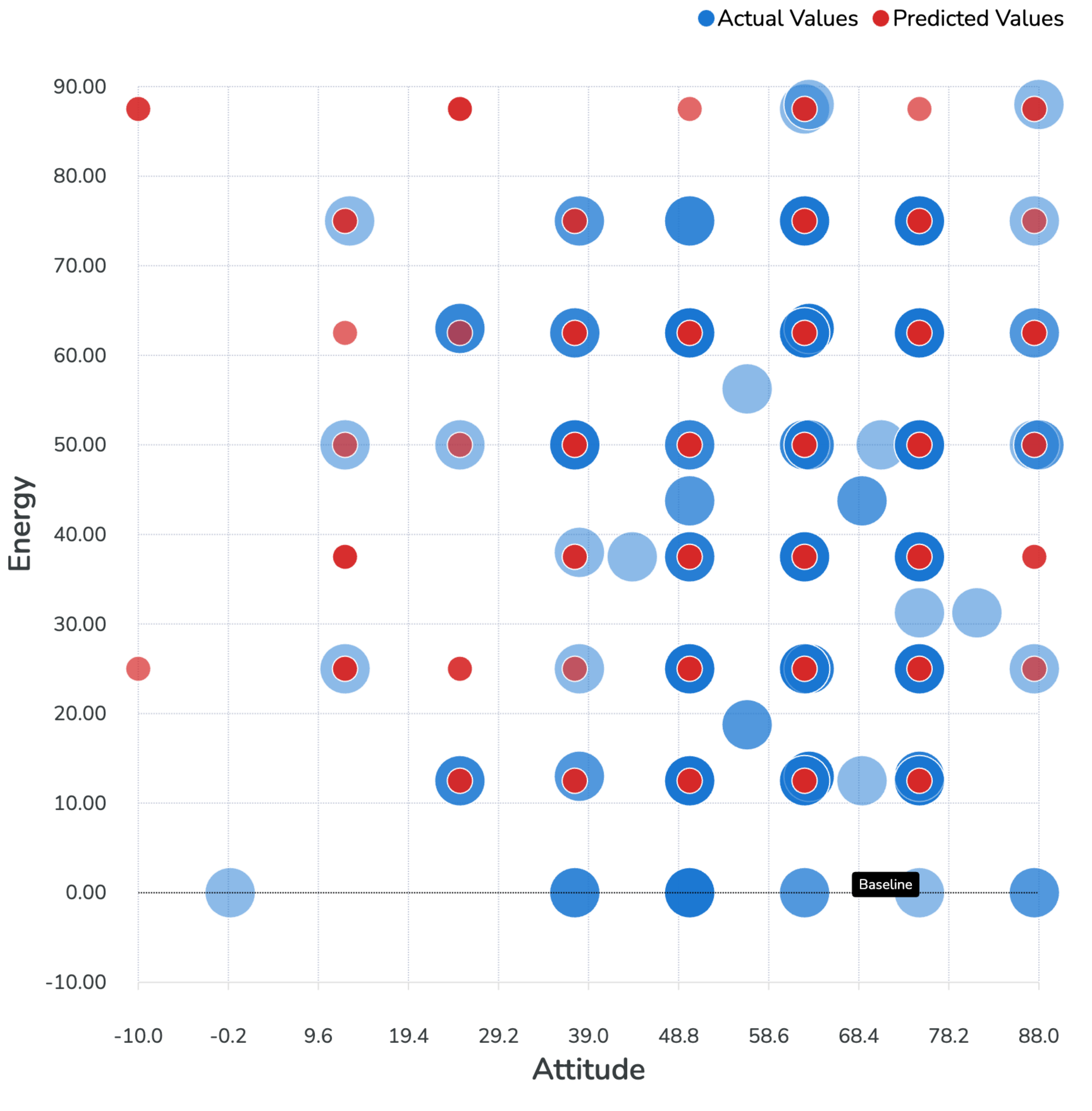}
        \caption{KNN}
        \label{fig:knn-metrics}
    \end{subfigure}
    \begin{subfigure}{0.325\linewidth}
        \centering
        \includegraphics[width=\linewidth]{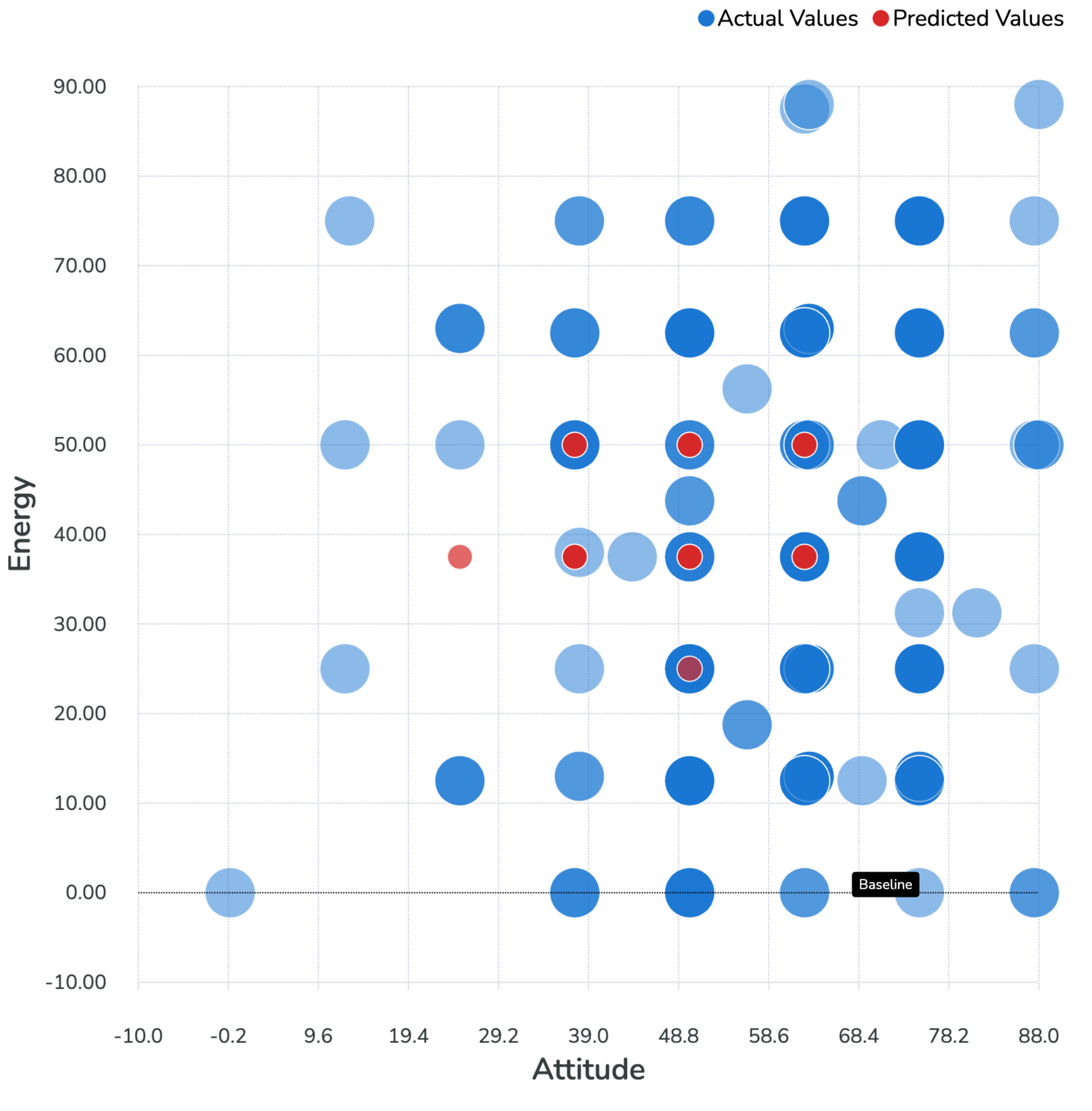}
        \caption{XGBoost}
        \label{fig:xgb-metrics}
    \end{subfigure}
    \begin{subfigure}{0.325\linewidth}
        \centering
        \includegraphics[width=\linewidth]{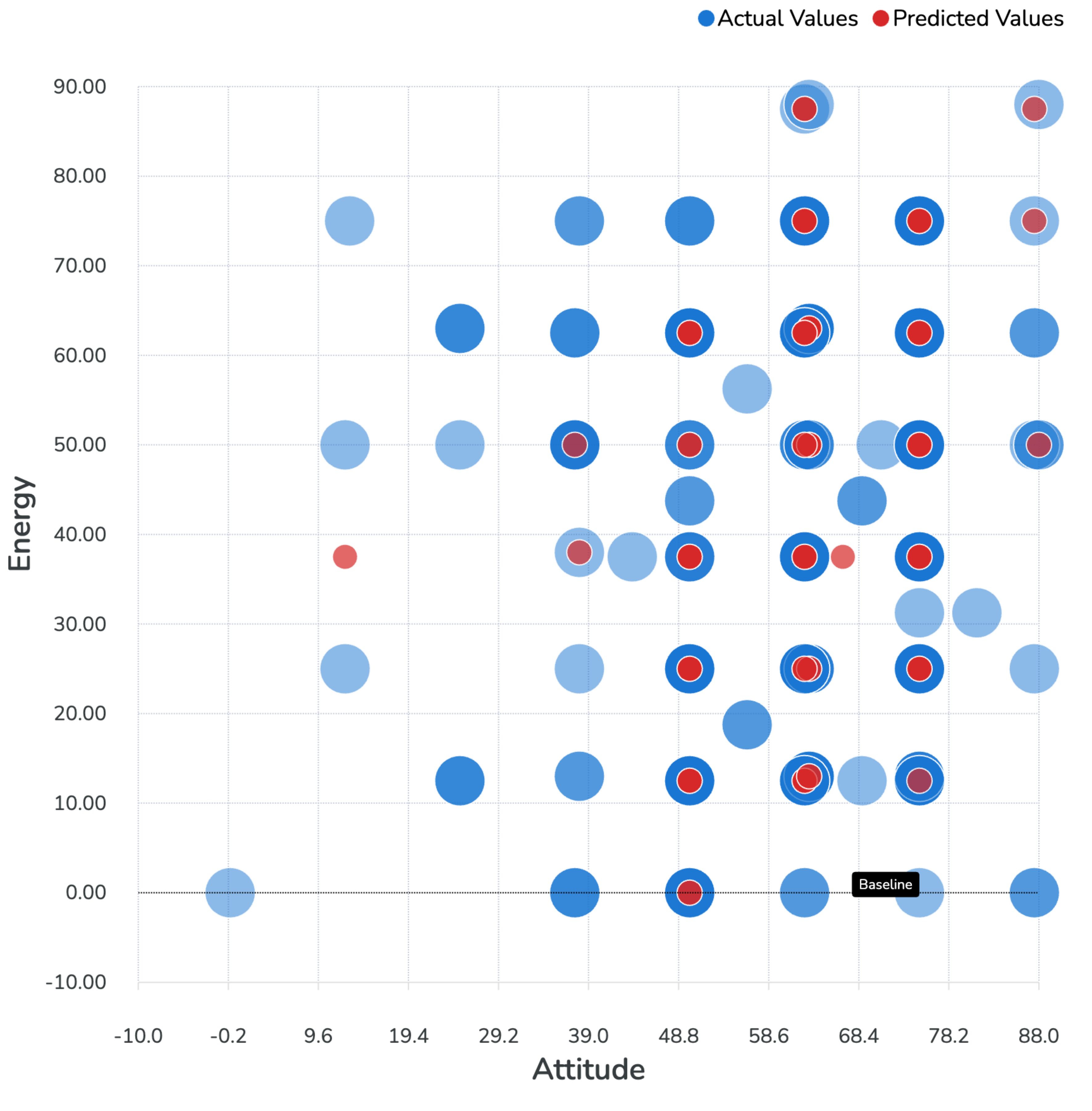}
        \caption{MSPSC}
        \label{fig:rlhf-metrics}
    \end{subfigure}
    \caption{Qualitative analysis of prediction qualities of candidate models on the test data. The plot matches emotion grid described in Figure~\ref{fig:emotion_grid}. KNN and XGBoost try to achieve some overlap of actual and predicted values but have considerable outliers and overfitting respectively. The blue circle shows actual user check-ins, whereas the red circle shows models' predictions.}
    \label{fig:comparative_analysis_actual_vs_predicted}
\end{figure*}

The general evaluation process involved splitting the dataset into one training set and five testing sets, using the test data to gauge the model’s generalization ability. This included both consistent and inconsistent users. Accuracy was determined by the absolute delta between predicted and actual emotion values, with predictions deemed correct if both deltas fell within a 13-unit threshold, aligned with UpBeing’s 64-point emotion grid. Table \ref{tab:comparative_analysis} summarises the performance of various models for consistent and inconsistent users. The better performance of MSPSC in in-consistent users suggests that MSPSC can generalize well even when user data is sporadic, making it robust in varied scenarios. Its adaptability is crucial for improving user experience by providing consistent emotional insights, regardless of the user’s interaction pattern.

We further performed a case study on one of the most promising user, in terms of consistent check-ins and access to data from environmental sources, to evaluate the performance of the aforementioned models. Table \ref{tab:comparative_analysis_single_user} summarizes the performance of each model for the selected consistent user. Figure \ref{fig:comparative_analysis_actual_vs_predicted} further illustrates the actual vs predicted outputs. The higher accuracy suggests that MSPSC is better at generalizing across different emotional states, making it more adaptable to the user's emotional shifts. The slightly higher Delta could indicate that MSPSC, while being more accurate overall, is also exploring a broader range of emotional predictions, thereby allowing for more personalized predictions. In contrast, other models may have lower Delta values but fail to capture the user's emotional patterns as effectively, resulting in lower accuracy.

\section{Discussion}
The Mood Shift Predictor with Sparse Check-ins (MSPSC) exemplifies a pioneering approach to addressing the complex, multifaceted nature of human emotions within AI systems. By integrating psychological understanding of emotional shifts with advanced machine learning techniques, MSPSC bridges the gap between basic AI research and real-world applications. This system is designed to overcome one of the most significant barriers in emotion prediction: data sparsity, particularly in environments reliant on user inputs.

Traditional AI models, which often depend on large datasets and fixed feature sets, fall short in capturing the nuanced emotional states of individuals. These models struggle to adapt to the dynamic and personal nature of human emotions, limiting their ability to provide genuinely empathetic AI (EAI) experiences. MSPSC, however, leverages a probabilistic framework that continuously learns from limited data, allowing it to dynamically adjust and predict emotional shifts with improved accuracy. This capability aligns closely with the psychological understanding that emotions are not static but evolve based on various factors.

This research highlights the innovative application of AI to a real-world problem—understanding and predicting human emotions. The MSPSC approach not only addresses current challenges in AI deployment, such as data sparsity and personalization, but it also paves the way for future developments in AI systems designed to be more responsive and adaptive to individual users. By focusing on the psychological aspects of emotion and integrating these insights into machine learning models, this work demonstrates a clear trajectory toward practical deployment in applications where user engagement and emotional well-being are paramount.

When deployed as a feature in the UpBeing App, the MSPSC system has the potential to significantly enhance user engagement by providing insights into users' emotional states, enabling timely interventions and personalized experiences. It illustrates a clear path forward for deploying these technologies in real-world, user-centric environments. This research represents a crucial step in advancing AI from theoretical models to practical, empathetic tools that can make a tangible difference in people’s lives.

\section{Conclusion}

This paper presents the Mood Shift Predictor with Sparse Check-ins (MSPSC) powering UpBeing App, which addresses the challenge of predicting mood shifts—a critical task for mental health and personalized tools. Inspired by well developed psychological understanding of human emotion, the proposed solution combines probabilistic modeling and utility functions based on simulated user feedback to improve prediction accuracy by adapting to individual mood patterns. It leverages latest machine learning techniques in an innovative way. The empirical study shows that MSPSC effectively utilizes limited data to provide better predictions, addressing the issues of data sparsity.


\bibliography{aaai25}

\end{document}